\documentclass[11pt]{article}
\usepackage{acl2014}
\usepackage{times}
\makeatletter
\newcommand{\@BIBLABEL}{\@emptybiblabel}
\newcommand{\@emptybiblabel}[1]{}
\makeatother
\PassOptionsToPackage{hyphens}{url}\usepackage{hyperref}
\usepackage{amsmath}
\usepackage{latexsym}
\usepackage{comment}
\usepackage[pass,paperwidth=21cm,paperheight=29.7cm]{geometry}
\usepackage{color}
\usepackage{graphicx}
\usepackage{caption}
\usepackage{subcaption}
\usepackage{xspace}
\usepackage[clock]{ifsym}
\usepackage{mathabx}
\usepackage{multirow}
\makeatletter
\newif\if@restonecol
\makeatother

\usepackage[lined,algonl,boxed]{algorithm2e}
\usepackage{rotating}
\usepackage{listings}
\usepackage[autostyle]{csquotes}
\usepackage[small,compact]{titlesec}

\newif{\ifhidecomments}
\hidecommentsfalse

\ifhidecomments
   \newcommand{\llee}[1]{}
  \newcommand{\chenhao}[1]{}
\else
  \newcommand{\llee}[1]{\textcolor{red}{#1}}
 \newcommand{\chenhao}[1]{\textcolor{blue}{#1}}
\fi

\title{
A Corpus of Sentence-level Revisions in Academic Writing:\\
A Step towards Understanding Statement Strength in Communication
}

\author{Chenhao Tan\\
  Dept. of Computer Science \\
  Cornell University \\
  {\href{mailto:chenhao@cs.cornell.edu}{\tt chenhao@cs.cornell.edu}} \\\And
  Lillian Lee \\
  Dept. of Computer Science \\
  Cornell University \\
  {\href{mailto:llee@cs.cornell.edu}{\tt llee@cs.cornell.edu}} \\}

\begin{document}

\maketitle

\begin{abstract}
The strength with which a statement is made
 can have a significant impact on the audience.
For example, international relations can be strained by how the media in
one country describes an event in another;
and
papers can be rejected because
they overstate or understate
their
findings. It is thus important to understand the effects of
statement strength. A first step is to be able to distinguish between
strong and weak statements.  However, even this problem is understudied,
partly due to a lack of data.
Since strength is inherently relative, {\em revisions} of texts that
make claims are a natural
source of data on strength differences.
In this paper, we introduce
a corpus of sentence-level revisions from academic writing.
We also describe insights gained from our annotation
efforts for this task.
\end{abstract}

\section{Introduction}
\label{sec:intro}

\newcommand{\highlt}[1]{{\em #1}}
\begin{table*}[t]
\centering
\begin{tabular}{|l|p{6in}|}
\hline
ID & Pairs\\
\hline
\multirow{2}{*}{1} & S1: The algorithm is \highlt{studied} in this paper . \\
& S2: The algorithm is \highlt{proposed} in this paper .\\
\hline
\multirow{2}{*}{2} & S1: ... circadian pattern and burstiness in \highlt{human
communication activity} .
\\
& S2: ... circadian pattern and burstiness in \highlt{mobile
phone communication} .\\
\hline
\multirow{2}{*}{3} &
S1: ... using minhash techniques , \highlt{at a significantly} lower \highlt{cost and with same privacy guarantees} .\\
& S2: ... using minhash techniques , \highlt{with} lower \highlt{costs} .\\
\hline
\multirow{2}{*}{4} & S1: the rows and columns of the covariate matrix
\highlt{then} have \highlt{certain physical} meanings ...
\\
& S2: the rows and columns of the covariate matrix \highlt{could} have
 \highlt{different} meanings ...
\\
\hline
\multirow{2}{*}{5} & S1: they maximize the expected revenue of the
seller but \highlt{induce efficiency loss} .\\
& S2: they maximize the expected revenue of the seller but \highlt{are inefficient} .\\
\hline

\hline
\end{tabular}
\caption{Examples of potential strength differences.
\label{tb:intro}}
\end{table*}

It is important for
authors and speakers to find the appropriate ``pitch'' to
convey a desired message to the public.
Indeed,
sometimes heated debates can
arise around the choice of statement strength. For instance, on March 1,
2014,
an attack
at Kunming's railway
station left 29 people dead and more than 140 others
injured.\footnote{\url{http://en.wikipedia.org/wiki/2014_Kunming_attack}}
In the aftermath, Chinese media accused Western media of ``soft-pedaling
the attack and failing to state clearly that it was an act of {\em
  terrorism}''.\footnote{\url{http://sinosphere.blogs.nytimes.com/2014/03/03/u-n-security-council-condemns-terrorist-attack-in-kunming/}}
In particular, regarding the statement by the US embassy that referred
to this incident as the ``terrible and
senseless act of violence in Kunming'', a Weibo user posted
``If you say that the Kunming attack is a `terrible and senseless act
of violence', then the 9/11 attack can be called a `regrettable
traffic
incident'''.\footnote{\url{http://www.huffingtonpost.co.uk/2014/03/03/china-kunming-911_n_4888748.html}%
}
This example is striking but not an isolated case, for
settings in which one party is trying to convince another are
pervasive;
scenarios range from court trials to conference
submissions.
Since the strength and scope of an
argument can be a crucial factor in its success,
it is
important to understand the effects of
statement strength in communication.

A first step towards addressing this question is to be able to
distinguish between strong and weak statements.
As strength is inherently
relative, it is natural to look at
{\em revisions} that change
statement strength, which we
refer to as ``{\em strength changes}''.
Though careful and repeated revisions are presumably ubiquitous in
politics, legal systems, and journalism, it is
not clear how to
collect
them;
on the other hand,
revisions to research papers may be more accessible, and
many researchers spend significant time
on editing
to
convey the right message regarding the strength of a project's contributions, novelty, and limitations. Indeed,
statement strength in science communication matters
to writers:
understating contributions can affect whether people
recognize the true importance of the work;
at the same time, overclaiming can cause papers to be rejected.

With the increasing popularity of e-print services
such as the arXiv\footnote{\url{http://arxiv.org/}},
strength changes in scientific papers are becoming more
readily available.
Since the arXiv started
in 1991,
it has become ``the standard repository for new papers
in mathematics, physics, statistics, computer science, biology, and
other disciplines'' \cite{Krantz:2007}.
An intriguing observation is that many researchers submit multiple versions of
the same paper on arXiv. For instance, among the 70K
papers submitted in 2011, almost 40\% (27.7K)
have multiple versions.
Many
differences between these versions constitute a source of valid
and motivated strength differences, as can be seen from the sentential
revisions in Table \ref{tb:intro}.
Pair 1 makes the contribution seem more impressive by replacing
``studied'' with ``proposed''. Pair 2 downgrades ``human communication
activity'' to ``mobile phone communication''. Pair 3 removes
``significantly'' and the emphasis on ``same privacy guarantees''.
Pair 4 shows an insertion of hedging, a relatively well-known type of strength reduction.
Pair 5 is an interesting case that shows the complexity of this
problem: on the one hand, S2
claims that
 something is
``inefficient'', which is an absolute statement,
compared to ``efficiency loss'' in S1, where the possibility of
efficiency still exists; on the
other hand, S1
employs an
active tone that emphasizes a causal relationship.

The main contribution of this work is to provide the first large-scale
corpus of sentence-level revisions for studying
a broad range of variations in statement strength.
We collected labels for a subset of
these
revisions.
Given the
possibility of all kinds of disagreement, the fair level of agreement
(Fleiss' Kappa)
among our annotators
was decent. But
in some cases, the labels
differed from our expectations,
indicating
that the general public can interpret the strength of scientific
statements differently from researchers.
The participants' comments
may further
shed light on science communication and point to better ways to define and understand
strength differences.

\section{Related Work and Data}

\newcommand{\parcap}[1]{\parbox[t]{0.9\textwidth}{#1}}
\begin{figure*}[t]
\centering
\begin{subfigure}[t]{0.32\textwidth}
  \includegraphics[width=\textwidth]{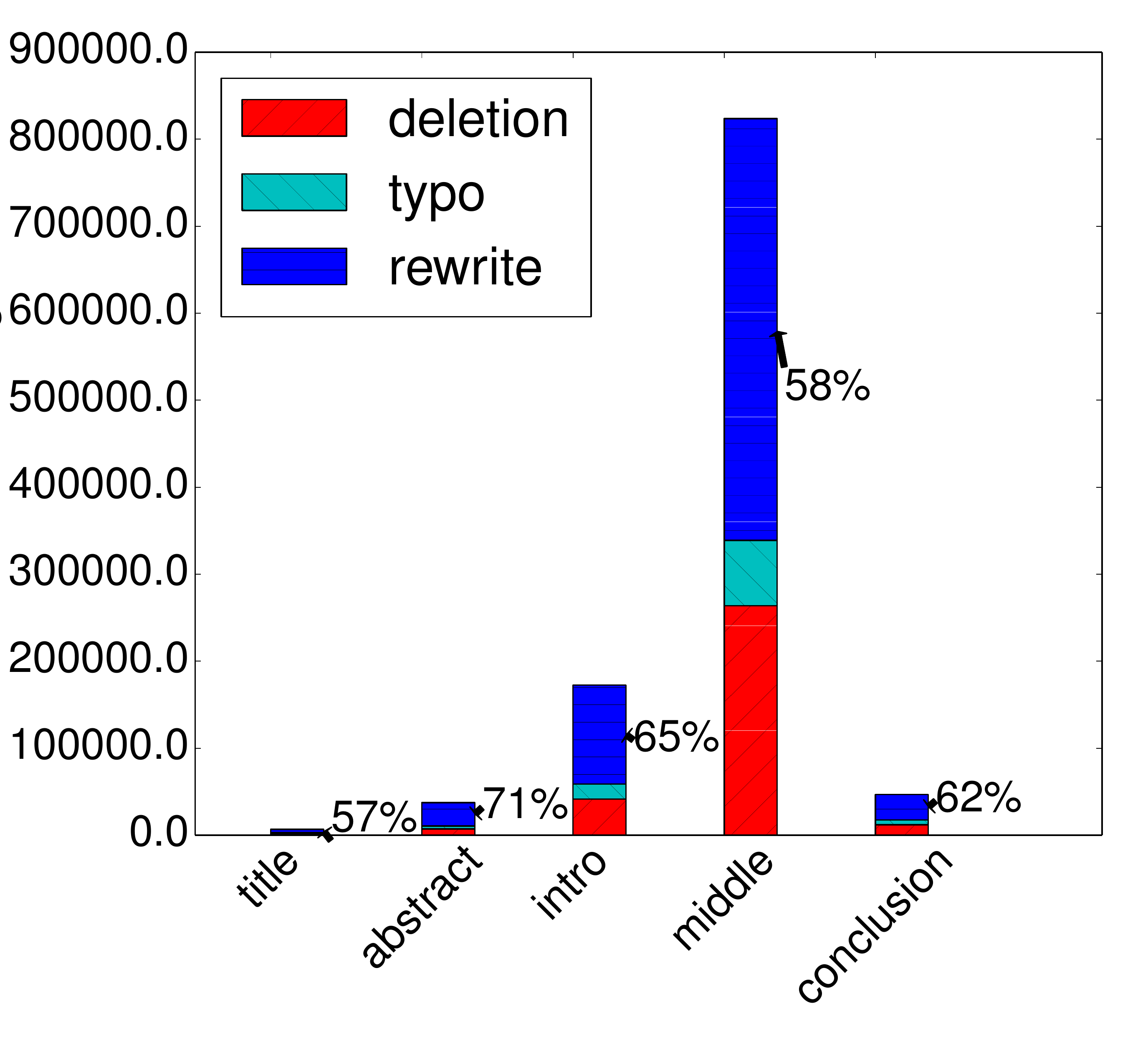}
  \caption{\parcap{Number of changes vs sections.
  ``middle'' refers to the sections between introduction and
  conclusion.}}
  \label{fig:section}
\end{subfigure}
\hfill
\begin{subfigure}[t]{0.32\textwidth}
  \includegraphics[width=\textwidth]{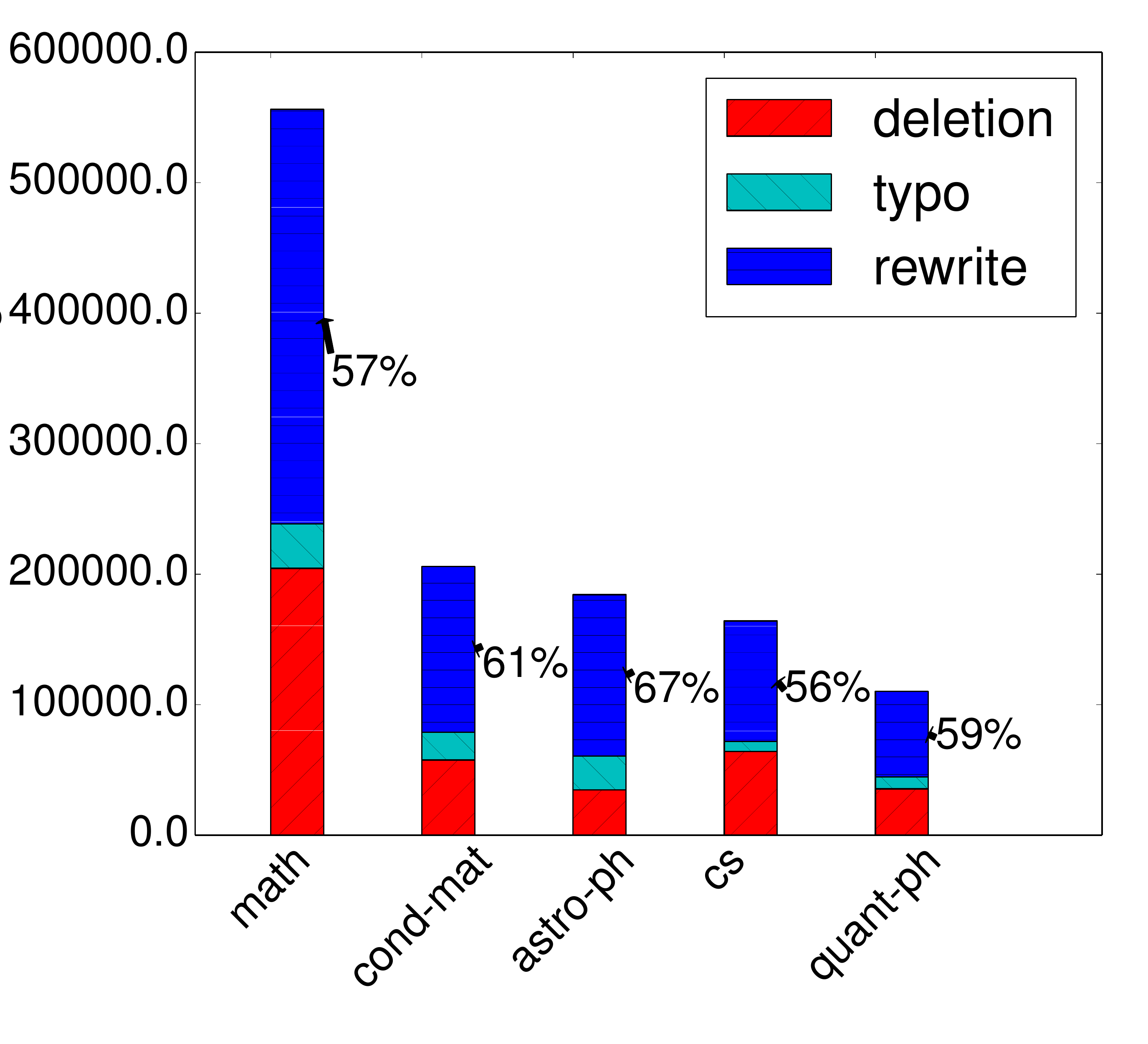}
  \caption{\parcap{Top 5 categories in number of changes.}}
  \label{fig:category_change}
\end{subfigure}
\hfill
\begin{subfigure}[t]{0.32\textwidth}
  \includegraphics[width=\textwidth]{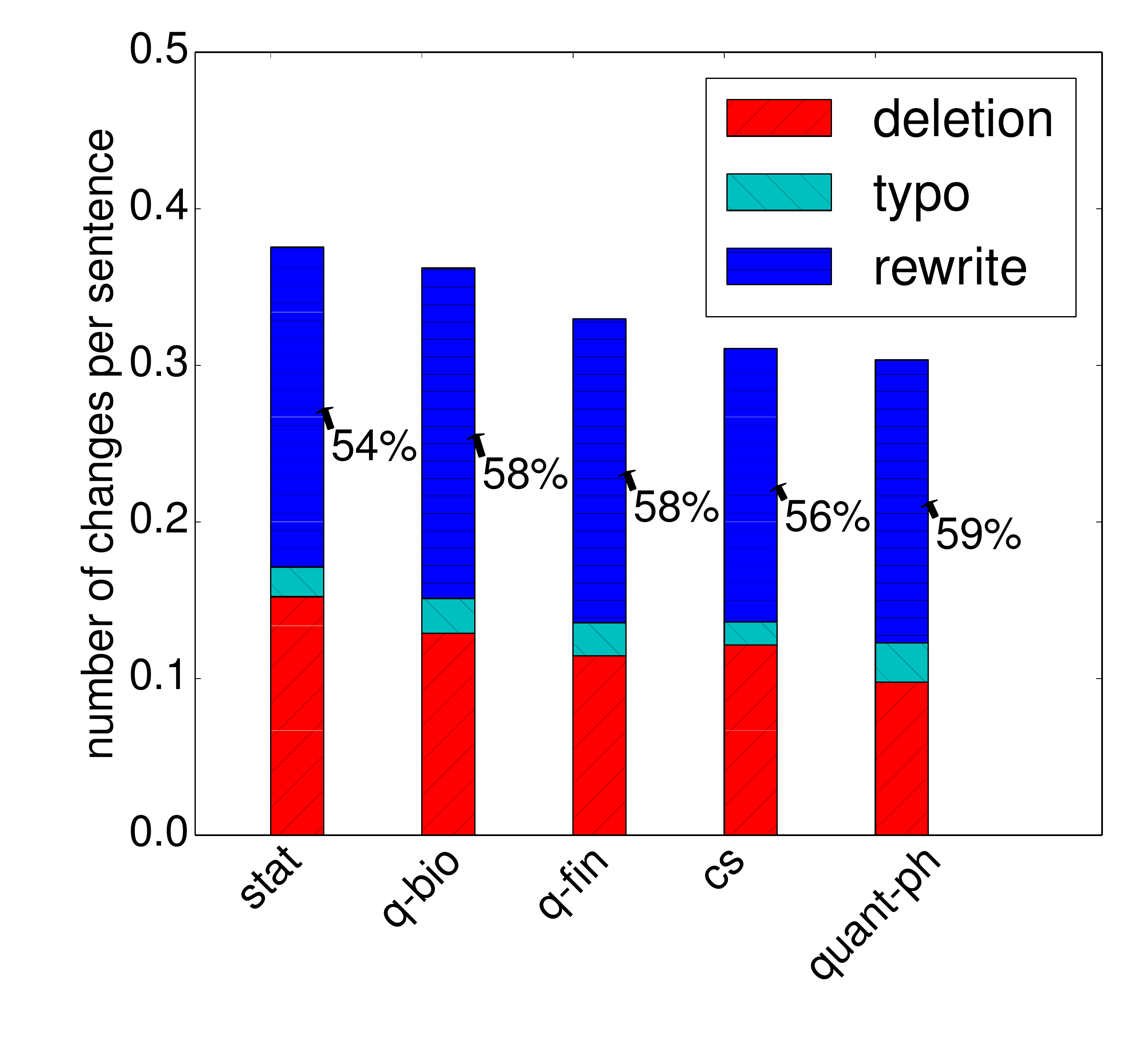}
  \caption{\parcap{Top 5 categories in number of changes over the number of
    sentences.}}
  \label{fig:category_percentage}
\end{subfigure}
\caption{In all figures, different colors
indicate different types of changes.
\label{fig:category}}
\end{figure*}

Hedging, which can lead to strength differences,
has received some attention in the study of science communication \cite{Salager-Meyer:2011a,Lewin:98a,Hyland:98a,Myers:90a}.
The CoNLL 2010 Shared Task was devoted to hedge detection \cite{Farkas+al:2010a}.
Hedge detection was also used to understand scientific framing in
debates over genetically-modified organisms in food
\cite{Choi+al:2012}.
Revisions on Wikipedia have been shown
useful for various applications, including
spelling correction \cite{Zesch:2012:MCF:2380816.2380880},
sentence compression \cite{Yamangil:2008:MWR:1557690.1557726}, text simplification \cite{yatskar2010sake},
paraphrasing \cite{max2010mining}, and textual entailment
\cite{zanzotto2010expanding}.
But none of the
categories of Wikipedia revisions
previously examined
\cite{daxenbergerautomatically,Bronner:2012:UEC:2380816.2380860,Mola-Velasco:2011:WVD:1963192.1963349,Potthast:AdvancesInInformationRetrieval:2008,daxenberger2012corpus}
relate to
statement strength. After all, the objective of editing on Wikipedia is to
present neutral and objective articles.

Public datasets of science communication are available, such as the ACL
Anthology,\footnote{\url{http://aclweb.org/anthology/}}
collections of
 NIPS papers,\footnote{\url{http://nips.djvuzone.org/txt.html}} and so on. These
datasets are useful for understanding the progress of
disciplines or the evolution of topics. But the lack of edit histories or
revisions makes them not
immediately suitable for studying strength
differences. Recently, there
have been
experiments with {\em open} peer review.\footnote{\url{http://openreview.net}}
Records from open reviewing
can provide additional insights
into the revision process
once enough data is
collected.

\section{Dataset Description}
\newcommand{\rewrite}{rewrite\xspace}
\newcommand{\Rewrite}{Rewrite\xspace}

Our main dataset was constructed from all
papers submitted in 2011 on the arXiv.
We first extracted
the
textual content from papers that have multiple
versions of tex source files. All
mathematical environments were ignored.
Section titles were not included in the final texts but are used in
 alignment.

In order to align the first version and the final version of the same
paper,
we first did macro alignment
of paper sections
 based on section
titles. %
Then, for micro alignment
of sentences,
we employed a dynamic programming
algorithm similar to
that of
\newcite{Barzilay:2003:SAM:1119355.1119359}.
Instead of cosine similarity, we used
an
idf-weighted longest-common-subsequence
algorithm
 to define the similarity between
two sentences,  because changes in word
ordering can also be interesting. Formally, the similarity score between
  sentence $i$ and sentence $j$ is defined as
\begin{equation*}
Sim(i,j)=\frac{\mbox{Weighted-LCS}(S_i, S_j)}{max(\sum_{w \in S_i}idf(w),\sum_{w
      \in S_j}idf(w))},
\end{equation*}
where $S_i$ and $S_j$ refer to
  sentence $i$ and sentence $j$.
Since it is likely that a new version adds or deletes a large
sequence of sentences, we did not impose
a skip penalty.
We set the mismatch penalty
to 0.1.\footnote{We did not allow cross matching (i.e., $i\rightarrow j-1$, $i-1
  \rightarrow j$), since we
  thought matching
this case as
$(i-1,i) \rightarrow j$ or $i
\rightarrow (j,j-1)$ can provide
context for annotation purposes.
But in the end, we focused on
labeling very similar pairs. This decision had little effect.}

In the end, there are 23K
papers where the first version was
different from the last version.\footnote{
This differs from the
  number in Section \ref{sec:intro} because articles may not have the tex
  source available, or the differences between versions may be in non-textual content.}
We categorize
sentential revisions
into the
following three types:
\begin{itemize}
\itemsep0em
\item Deletion: we cannot find a match in the final
  version.
\item Typo: all sequences in a pair of matched sentences are typos, where a sequence-level typo is one where the edit distance between the matched sequences is less than three.
\item \Rewrite:
matched sentences that are
  not typos. This type is
the focus of this study.

\end{itemize}

\begin{table*}
\centering
\begin{tabular}{|p{6in}|}
\hline
You should mark S2 as {\bf Stronger} if\\
$\bullet$ (R1) S2 strengthens the degree of some aspect of S1, for example, S1
  has the word "better", whereas S2 uses "best", or S2 removes the
  word "possibly"\\
$\bullet$ (R2) S2 adds more evidence or justification (we {\em don't} count
  adding details)\\
$\bullet$ (R3) S2 sounds more impressive in some other way: the authors' work
  is more important/novel/elegant/applicable/etc.\\
If instead S1 is stronger than S2 according to the reasons above,
select {\bf Weaker}.
If the changes aren't strengthenings or weakenings according to the reason above, select No Strength Change.\\
If there are both strengthenings and weakenings, or you find that it
is really hard to tell whether the change is stronger or weaker, then
select I can't tell.
\\
\hline
\end{tabular}
\caption{Definition of labels in our labeling tasks.\label{tb:defs}}
\end{table*}

\paragraph{What kinds of changes are being made?}
One
might initially
think that
typo fixes represent a large proportion of revisions,
but this is not correct, as shown in Figure \ref{fig:section}.
Deletions represent a substantial fraction, especially in
the middle section of a paper.
But it is clear
that the majority of changes are {\rewrite}s; %
thus revisions on the arXiv indeed provide a
great source for potential strength differences.
\paragraph{Who makes changes?}
Figure
\ref{fig:category_change} shows that the Math subarchive makes the largest number
of changes. This is consistent with
the mathematics community's custom of using the arXiv to get findings out early.
In terms of
changes per sentence (Figure \ref{fig:category_percentage}), statistics and quantitative studies are the top
subareas. %

\begin{figure}[t]
\centering
\begin{subfigure}[t]{0.22\textwidth}
  \includegraphics[width=\textwidth]{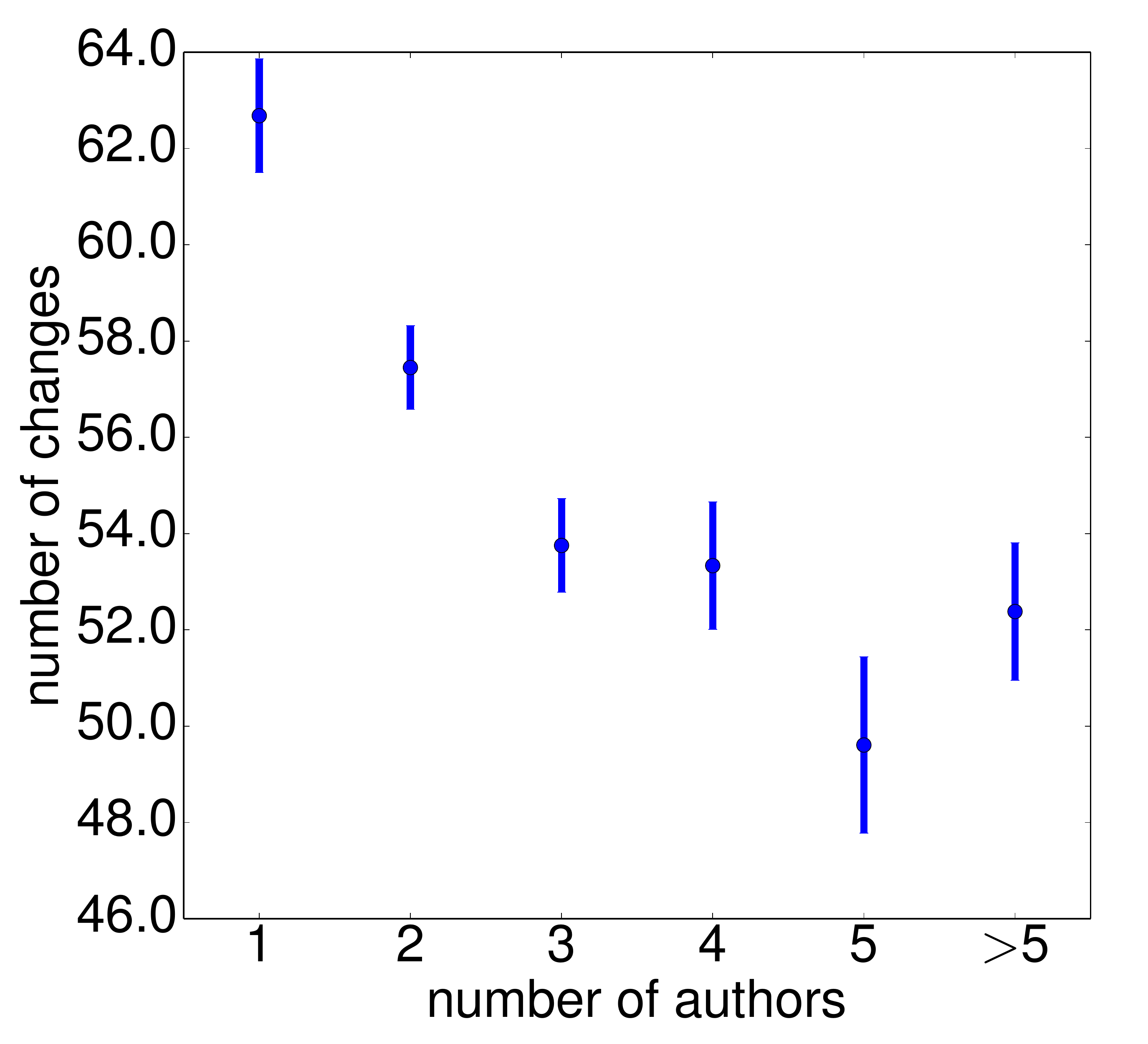}
  \caption{Number of changes vs number of authors.}
  \label{fig:author_change}
\end{subfigure}
\hfill
\begin{subfigure}[t]{0.22\textwidth}
  \includegraphics[width=\textwidth]{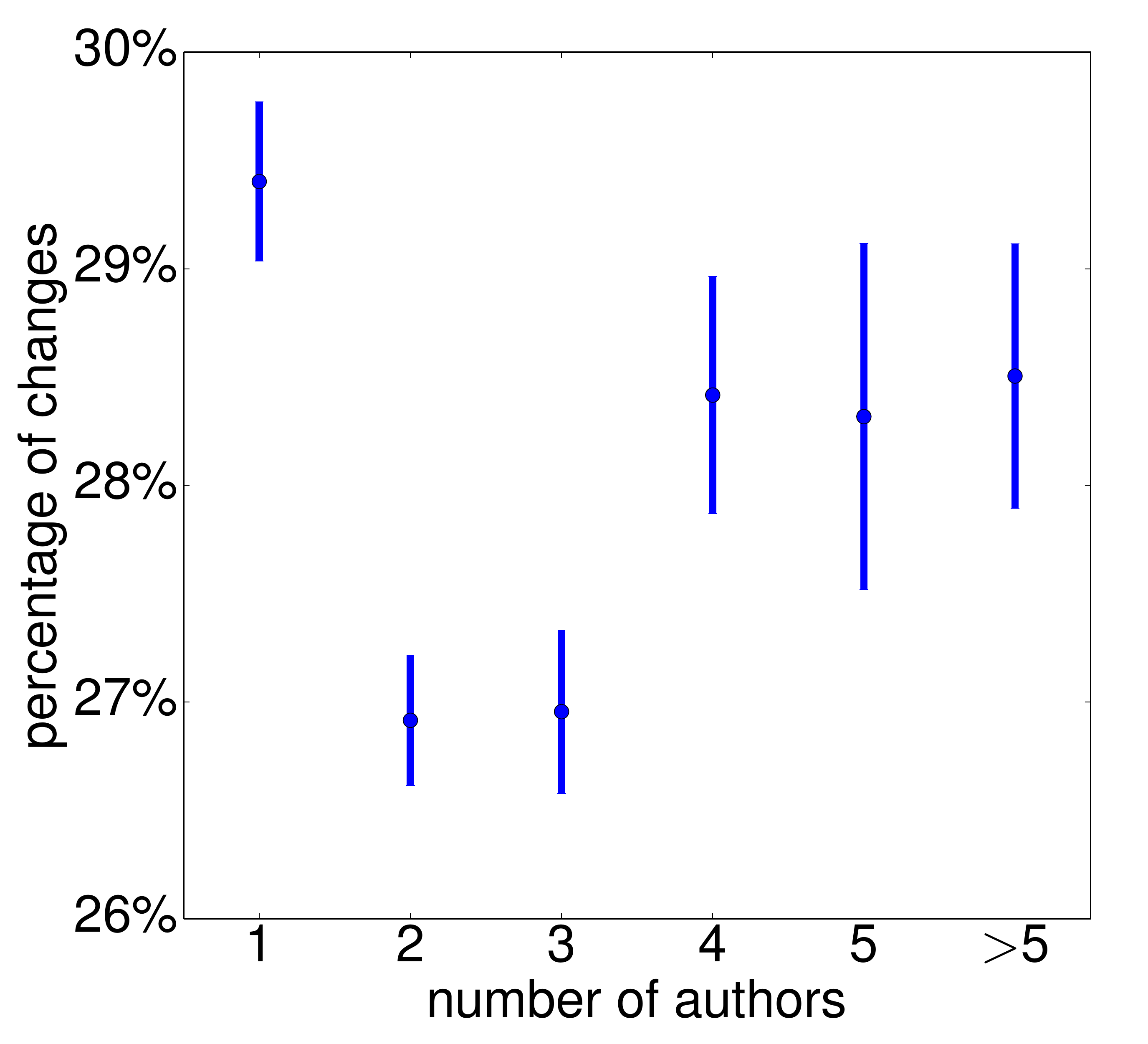}
  \caption{Percentage of changed sentences vs
number of authors.}
  \label{fig:author_percentage}
\end{subfigure}
\caption{Error bars represent standard error.
(a): up until 5 authors, a larger number of authors indicates a
smaller number of changes.
(b): percentage is measured over the number of sentences in the first
version; there is an interior minimum where 2 or 3 authors make the
smallest percentage of sentence changes on a paper.
\label{fig:author}}
\end{figure}

Further, Figure \ref{fig:author} shows the
effect of the number of
authors. It is interesting that both in terms of sheer
number and percentage, single-authored papers have the most
changes. This %
could be because a single author enjoys
greater freedom
and has stronger motivation to make changes,
or because multiple authors
tend to submit a more polished initial version. This echoes the finding in
\newcite{posner1992people} that the collaborative writing process differs considerably
from individual writing.
Also, more than 25\% of the first
versions
are changed, which again shows that
substantive
 edits are being made in
these resubmissions.

\section{Annotating Strength Differences}

In order to
study statement strength,
reliable strength-difference labels
are needed.
In this section, we describe how
we tried to define
strength differences, compiled labeling
instructions, and
gathered labels
using Amazon Mechanical Turk.

\begin{table*}
\centering
\small
\begin{tabular}{|l|p{6in}|}
\hline
ID & Matched sentences and comments\\
\hline
\multirow{4}{*}{1} & S1:
... using data from numerics and experiments .
\\
& S2: 
... using data sets from numerics in the point particle limit and one experimental data set .
\\
& (stronger) S2 is more specific in its description which seems stronger.
\\
& (weaker) "one experimental data set" weakens the sentence
\\
\hline
\multirow{4}{*}{2}
& S1: we also proved that if [MATH] is sufficiently homogeneous then ...
\\
& S2: we also proved that if [MATH] is not totally disconnected and sufficiently homogeneous then ...
\\
& (stronger) We have more detail/proof in S2
\\
& (stronger) the words "not totally disconnected" made the sentence sound more impressive.
\\
\hline
\multirow{4}{*}{3}
& S1: we also show in general that vectors of products of jack vertex operators form a basis of symmetric functions .
\\
& S2: we also show in general that the images of products of jack vertex operators form a basis of symmetric functions .
\\
& (weaker) Vectors sounds more impressive than images
\\
& (weaker) sentence one is more specific
\\
\hline
\multirow{4}{*}{4}
& S1: in the current paper we discover several variants of qd algorithms for quasiseparable matrices .
\\
& S2: in the current paper we adapt several variants of qd algorithms to quasiseparable matrices .
\\
& (stronger) in S2 Adapt is stronger than just the word discover. adapt implies more of a proactive measure.
\\
& (stronger) s2 sounds as if they're doing something with specifics already, rather than hunting for a way to do it
\\
\hline

\end{tabular}
\normalsize
\caption{Representative examples of surprising labels, together with
  selected labeler comments.\label{tb:samples}}
\end{table*}

\paragraph{Label definition and collection procedure.}
We focused on matched sentences
from abstracts
and introductions to maximize the proportion of strength differences
(as opposed to factual/no strength changes).
We required pairs to
have similarity score larger than 0.5 in our
labeling task to make pairs more comparable.
We also replaced all math environments with ``[MATH]''.\footnote{%
These decisions were made based on the results and feedback that we
got from graduate students in an initial labeling.} We obtained 108K
pairs that satisfy the above conditions,
available at
\url{http://chenhaot.com/pages/statement-strength.html}.
To %
create the pool of pairs for labeling, we randomly sampled 1000 pairs
and then removed pairs that we thought were processing errors.

We used Amazon Mechanical Turk.
It may initially seem surprising to have annotations of technical
statements not done by
domain experts; we did this intentionally because it is common to
communicate unfamiliar topics to the public in political and science
communication (we comment
on non-expert
rationales later).
We use the following set of labels: {\em
  Stronger, Weaker, No Strength Change, I can't tell}.
Table \ref{tb:defs} gives our
definitions.
The instructions included 8 pairs as examples and 10 pairs to label
as a
training exercise.
Participants were then asked to choose labels and write
mandatory comments for 50 pairs.
According to the comments written by
participants,
we believe that they did the labeling in good faith.

\paragraph{Quantitative overview.}
We collected 9 labels each for 500 pairs.
Among the 500 pairs, Fleiss' Kappa was 0.242, which indicates fair
agreement \cite{Landis+Koch:77}.
We took a conservative approach and only considered pairs with
an absolute majority label, i.e., at least 5 of 9 labelers chose the
same label. There are 386 pairs that satisfy this requirement (93
weaker, 194 stronger, 99 no change).
On this subset of pairs, Fleiss' Kappa is 0.322, and
74.4\% of pairs
were strength changes.
Considering all the possible disagreement,
this result was acceptable.

\paragraph{Qualitative observations.}

We were excited about the labels from these participants:
despite the apparent difficulty of the task, we found that many labels
for the 386 pairs were reasonable.
However, in some cases, the labels were counter-intuitive.
Table \ref{tb:samples} shows some representative examples.

First, participants tend to take details as evidence even when these
details are not germane to the statement.
For pair 1,
while one turker pointed out the decline in number of
experiments, most turkers simply
labeled it as stronger because it was more
specific. ``Specific'' turned out to be a common reason used
  in the comments, even though we said in the instructions that only additional justification
  and evidence matter.
This echoes the finding in
\newcite{Bell:JournalOfPersonalityAndSocialPsychology:1989} that even
unrelated details influenced judgments of guilt.

Second, participants interpret constraints/conditions
not in strictly logical ways,
seeming
to care
little about
scope
at times.
For instance, the majority labeled
pair 2
as
``stronger''.
But in S2
for that pair, the
result
holds for
strictly
fewer possible worlds.
But it should be said
that there are cases that labelers interpreted logically, e.g., ``compelling evidence''
subsumes
``compelling
  experimental evidence''.

Both of the above cases share the property that they seem to be
correlated with a tendency to judge lengthier statements as stronger.
Another interesting case that does not share this characteristic is that
participants can have a different understanding of
domain-specific terms. For pair 3, the majority thought that
``vectors'' sounds more impressive than ``images''; for pair 4, the
majority considered ``adapt'' stronger than ``discover''.
This issue is common when communicating new topics to the public not
only in science communication but also in politics and other
scenarios.
It may partly explain
miscommunications and misinterpretations of scientific studies in
journalism.\footnote{\url{http://www.phdcomics.com/comics/archive.php?comicid=1174}}

\section{Looking ahead}

Our observations regarding the annotation results raise questions regarding what is
a generalizable way to define strength differences, how to use the
labels that we collected,
and how to collect labels in the future.
We believe that this corpus of sentence-level revisions, together with
the labels and comments from participants, can
provide insights
into better ways to
approach this problem and help further understand strength of
statements.

One interesting direction that this enables is a potentially new kind
of learning problem. The comments indicate features that humans think
salient. Is it possible to automatically learn new features from the comments?

The ultimate goal of our study is to understand the effects of
statement strength
on the public, which can lead to
various applications in public communication.
\section*{Acknowledgments}
\newcommand{\finit}[2]{#1.}

We thank
\finit{J}{ason} Baldridge, 
\finit{J}{ordan} Boyd-Graber,
\finit{C}{hris} Callison-Burch,
and the reviewers
for helpful comments;
\finit{P}{aul} Ginsparg for providing data;
and 
\finit{S}{huo} Chen,
\finit{E}{lisavet} Kozyri,
\finit{M}{oontae} Lee,
\finit{I}{an} Lenz,
\finit{M}{yle} Ott,
\finit{J}{on} Park,
\finit{K}{arthik} Raman,
\finit{M}{ark} Reitblatt,
\finit{S}{udip} Roy,
\finit{A}{mit} Sharma,
\finit{R}{uben} Sipos,
\finit{A}{dith} Swaminathan,
\finit{L}{u} Wang,
\finit{W}{enlei} Xie,
\finit{B}{ishan} Yang
and the anonymous annotators for all their labeling help.
This work was supported in part by  NSF grant IIS-0910664 and a Google Research Grant.

\bibliographystyle{acl}
\bibliography{ref}
\end{document}